\documentclass[10pt,twocolumn,letterpaper]{article}

\usepackage{cvpr} % uncomment this for the final submission
\usepackage{times}
\usepackage{epsfig}
\usepackage{graphicx}
\usepackage{amsmath}
\usepackage{amssymb}

\usepackage{hyperref}

\usepackage{booktabs}
\usepackage{caption}
\usepackage{longtable} % Optional for splitting across pages
\usepackage{lscape} % Optional if you want landscape table

\usepackage{multirow}
\usepackage{geometry}
\begin{document}

%%%%%%%%% TITLE
\title{Evaluating Deep Learning-Based Face Recognition for Infants and Toddlers: Impact of Age Across Developmental Stages}

\author{Afzal Hossain\\
Clarkson University\\
New York, USA\\
{\tt\small afhossa@clarkson.edu}
% For a paper whose authors are all at the same institution,
% omit the following lines up until the closing ``}''.
% Additional authors and addresses can be added with ``\and'',
% just like the second author.
% To save space, use either the email address or home page, not both
\and
Mst Rumana Sumi\\
Clarkson University\\
New York, USA\\
{\tt\small sumima@clarkson.edu}
\and
Stephanie Schuckers\\
University of North Carolina at Charlotte\\
North Carolina, USA\\
{\tt\small sschucke@charlotte.edu}
}

\maketitle
\thispagestyle{empty}

%%%%%%%%% ABSTRACT
\begin{abstract}
   Face recognition for infants and toddlers presents unique challenges due to rapid facial morphology changes, high inter-class similarity, and limited dataset availability. This study evaluates the performance of four deep learning-based face recognition models FaceNet, ArcFace, MagFace, and CosFace on a newly developed longitudinal dataset collected over a 24 month period in seven sessions involving children aged 0 to 3 years. Our analysis examines recognition accuracy across developmental stages, showing that the True Accept Rate (TAR) is only 30.7\% at 0.1\% False Accept Rate (FAR) for infants aged 0 to 6 months, due to unstable facial features. Performance improves significantly in older children, reaching 64.7\% TAR at 0.1\% FAR in the 2.5 to 3 year age group. We also evaluate verification performance over different time intervals, revealing that shorter time gaps result in higher accuracy due to reduced embedding drift. To mitigate this drift, we apply a Domain Adversarial Neural Network (DANN) approach that improves TAR by over 12\%, yielding features that are more temporally stable and generalizable. These findings are critical for building biometric systems that function reliably over time in smart city applications such as public healthcare, child safety, and digital identity services. The challenges observed in early age groups highlight the importance of future research on privacy preserving biometric authentication systems that can address temporal variability, particularly in secure and regulated urban environments where child verification is essential.
\end{abstract}

%%%%%%%%%%%%%%%%%%%%%%%%%%%%%%%%%%%%%%%%%%%%%%%%%%%%%%%%%%%%%%%%%%%%%%%%%%%%%%%%%%%%%%%%%%%%%%%

\begin{figure*}[!t]
    \centering
    \begin{tabular}{c c c c c c c}
        \includegraphics[width=0.12\textwidth]{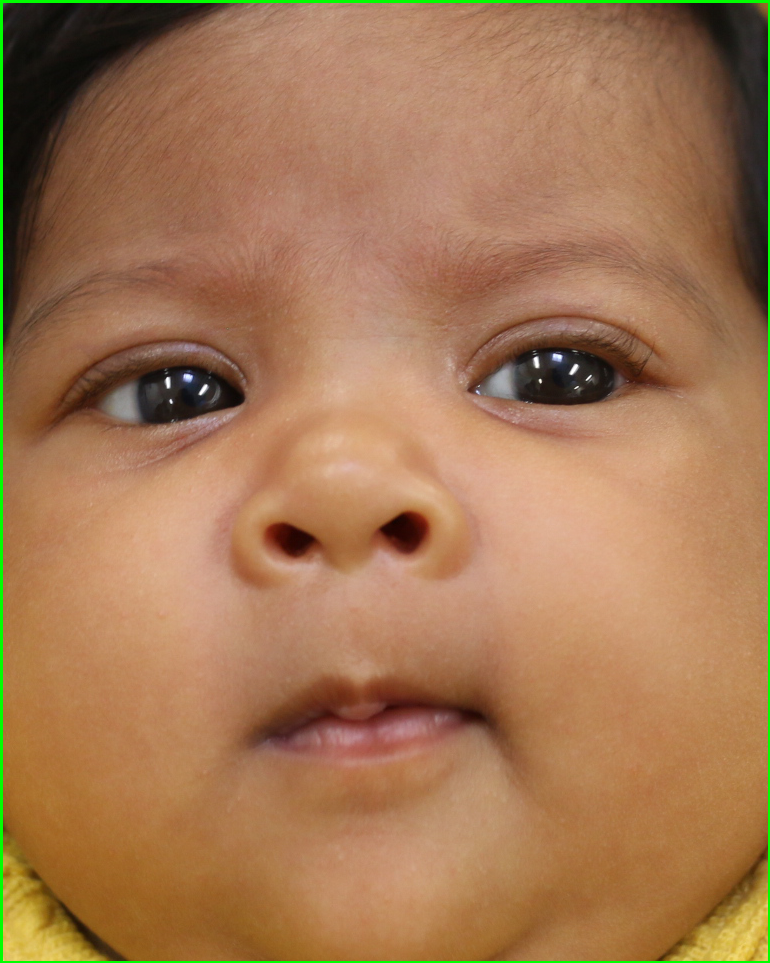} &
        \includegraphics[width=0.12\textwidth]{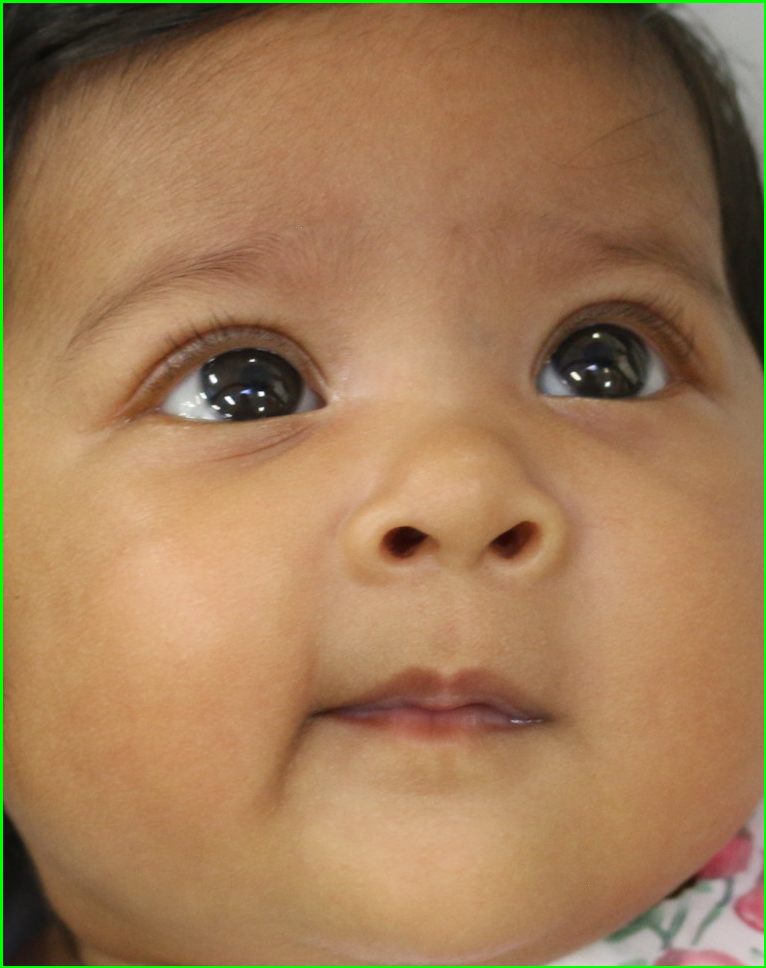} &
        \includegraphics[width=0.12\textwidth]{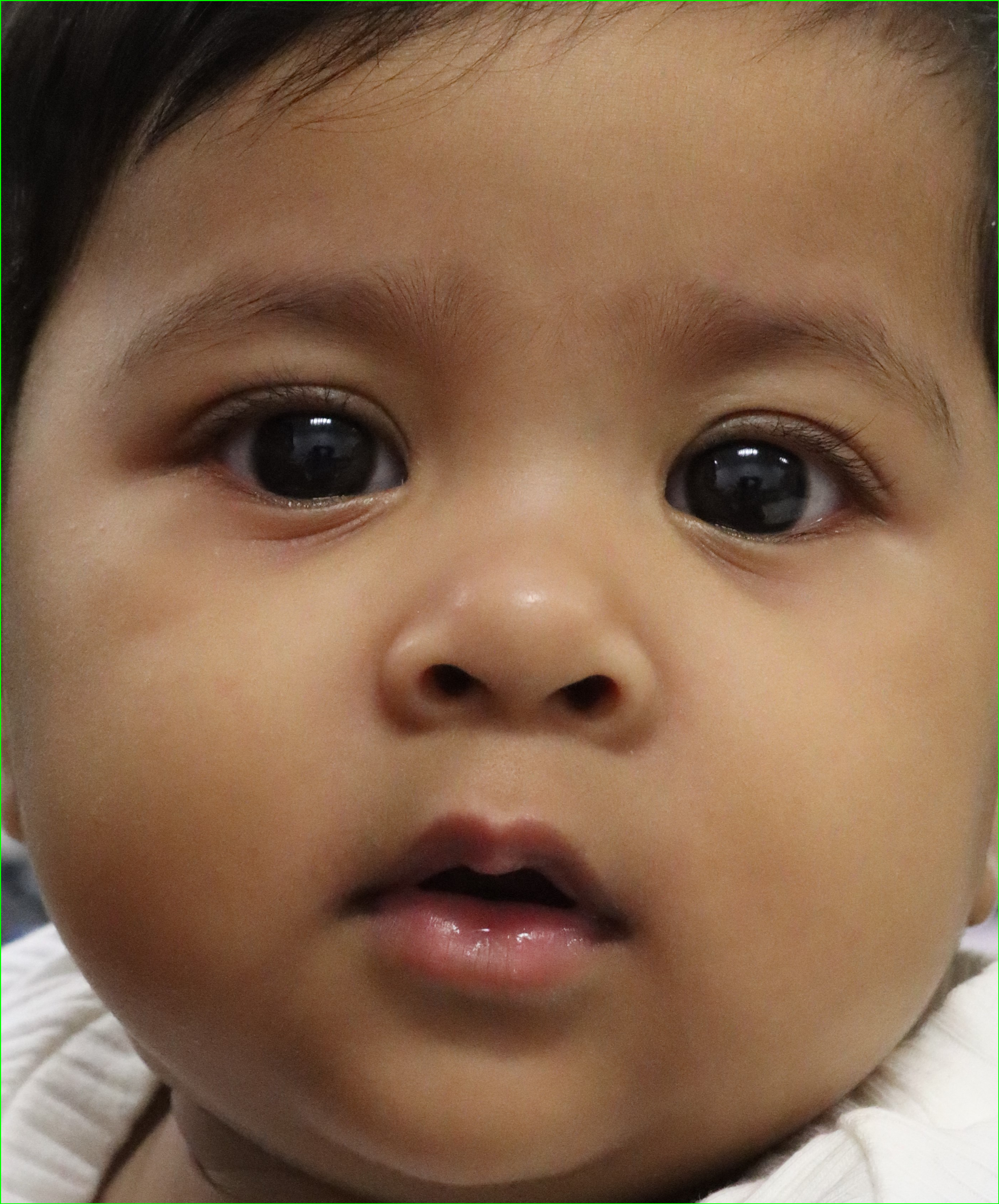} &
        \includegraphics[width=0.12\textwidth]{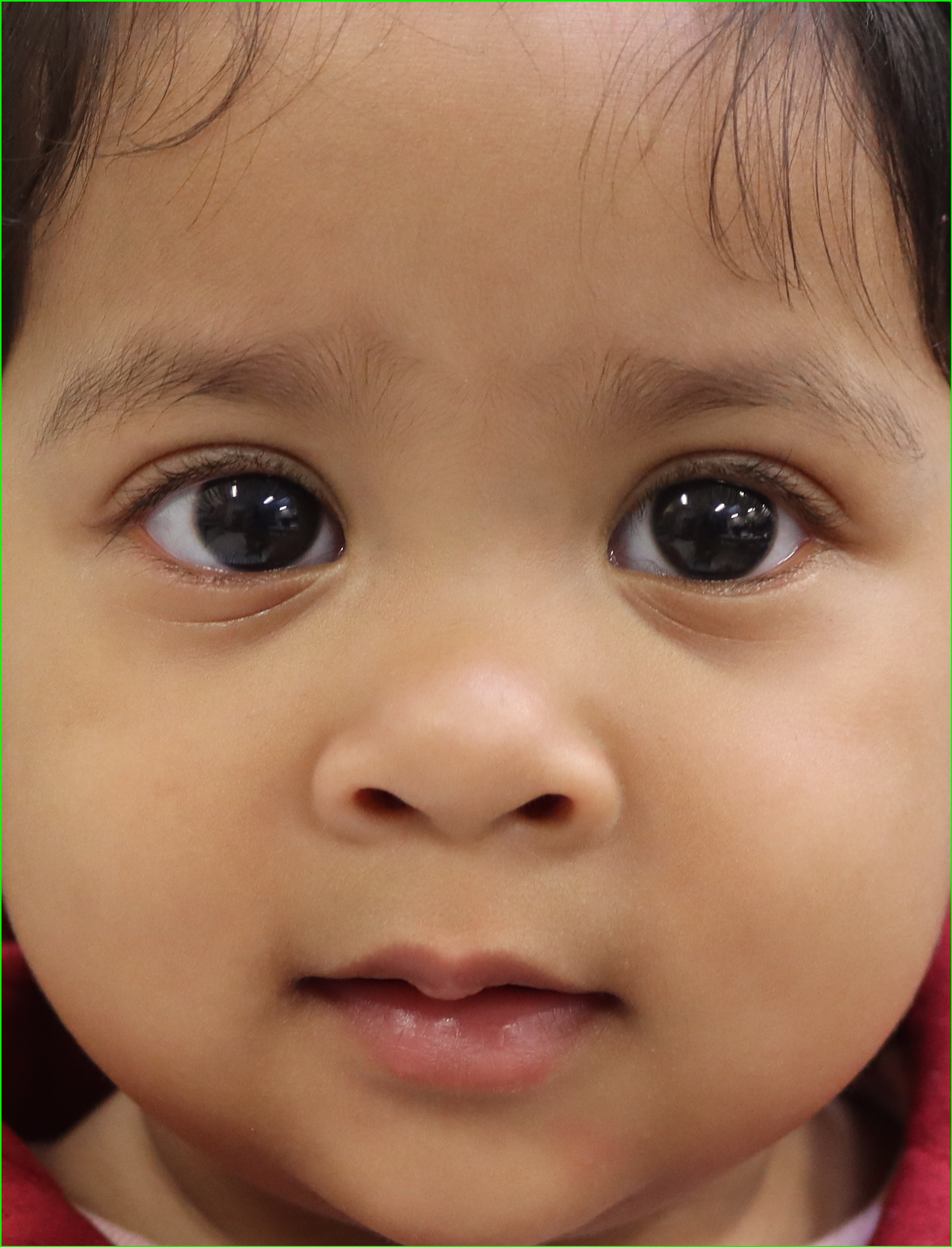} &
        \includegraphics[width=0.12\textwidth]{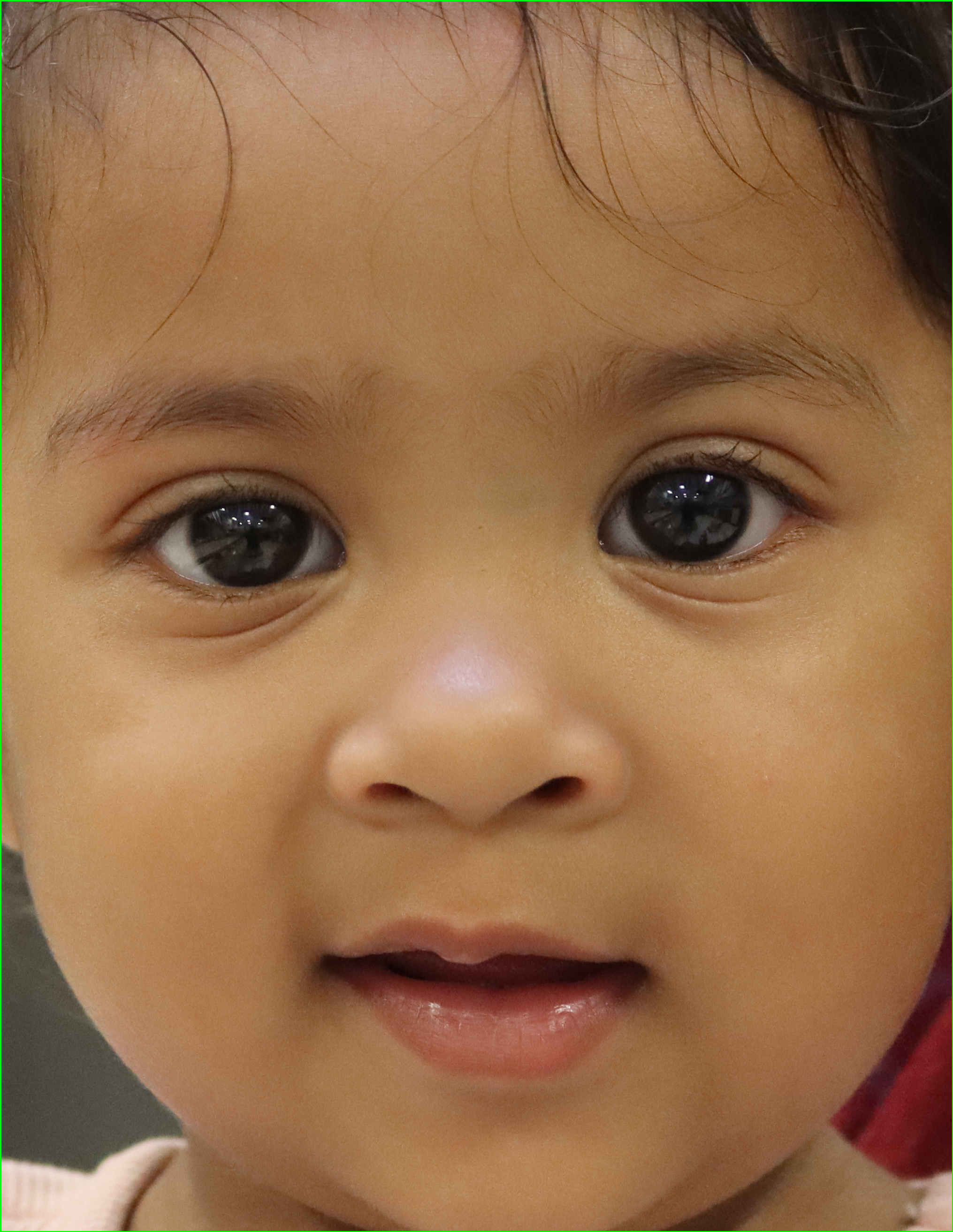} &
        \includegraphics[width=0.12\textwidth]{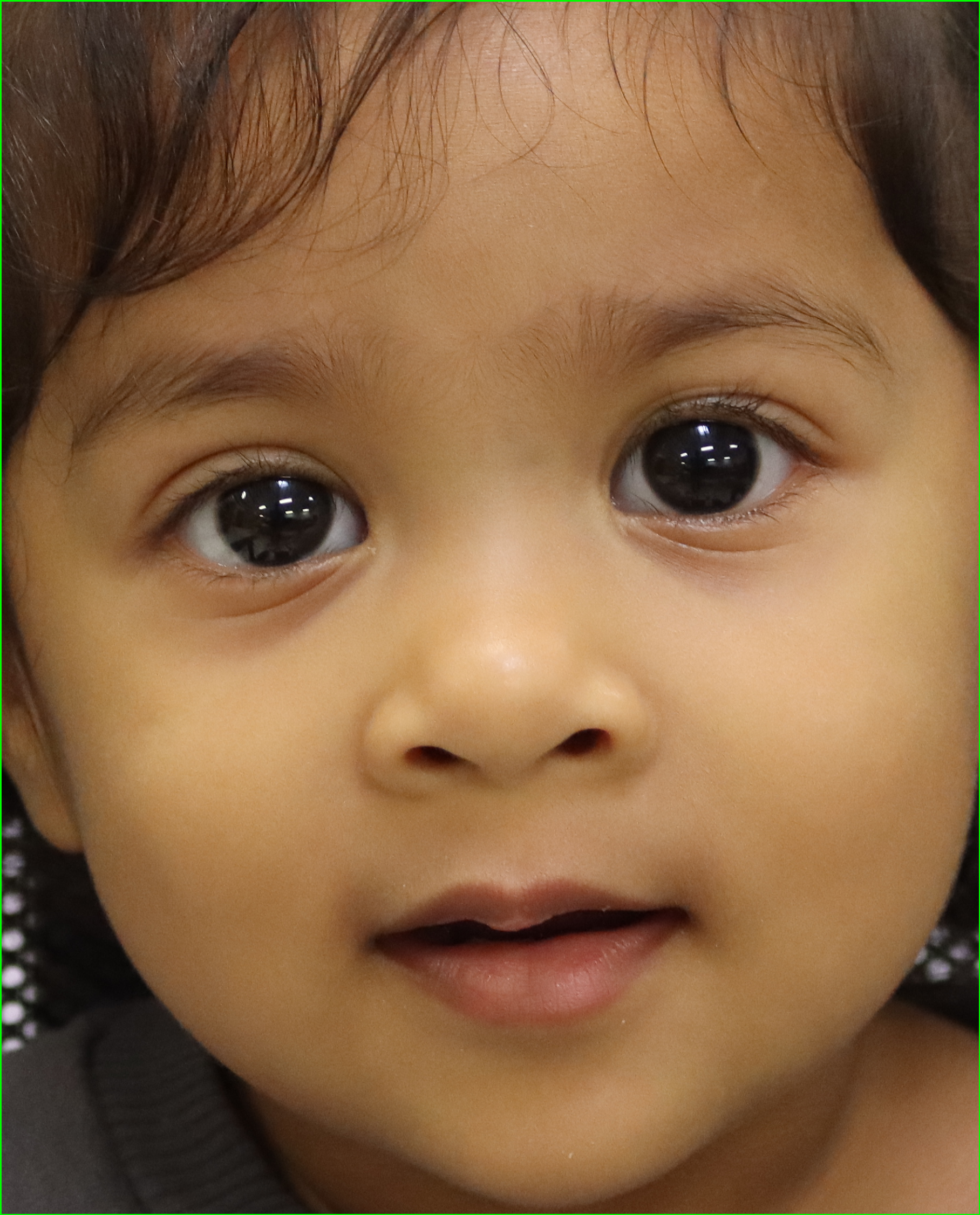} &
        \includegraphics[width=0.12\textwidth]{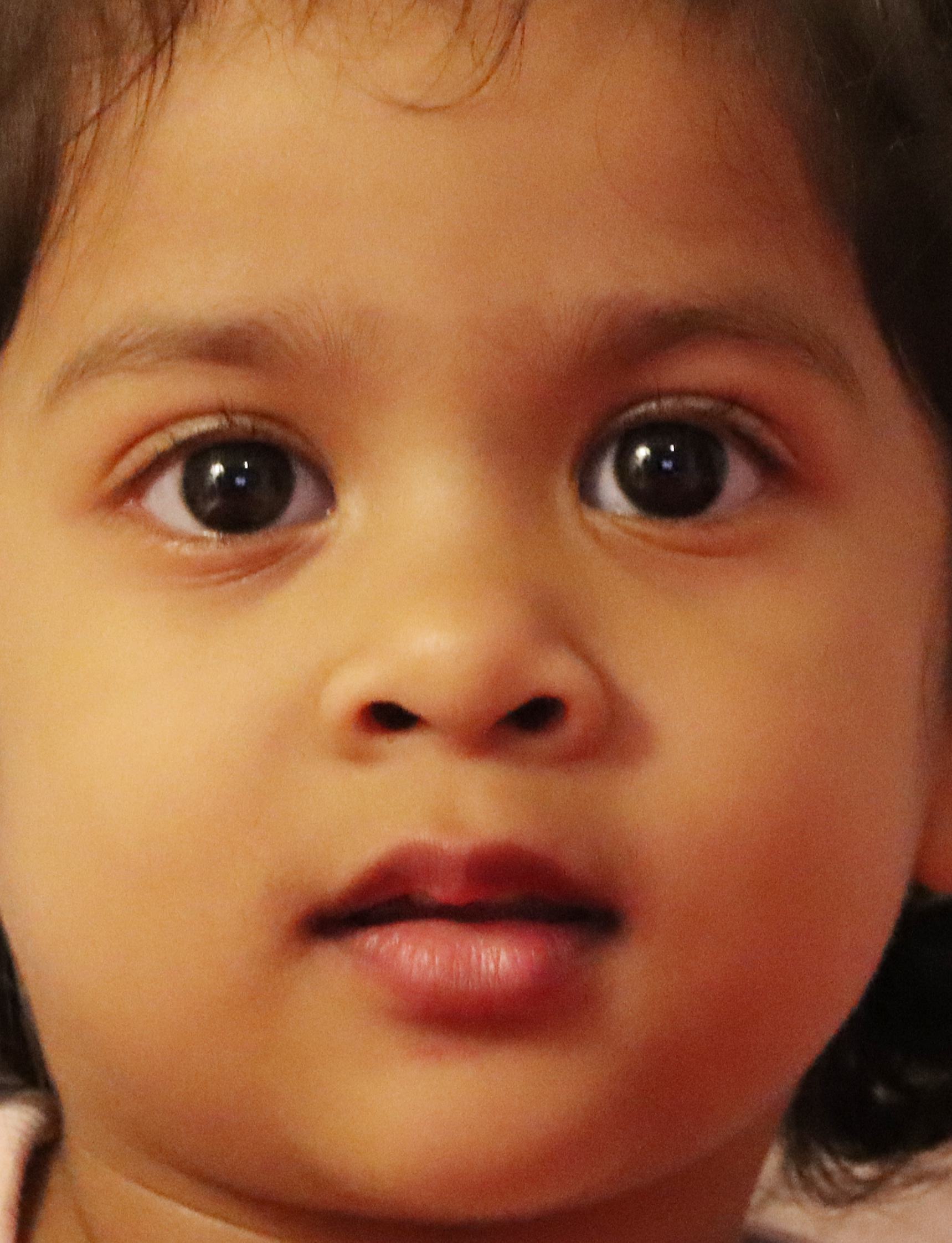} \\
        1 month & 2 months & 7 months & 11 months & 14 months & 18 months & 24 months \\
    \end{tabular}
    \caption{Age progression of a subject across different sessions in the Infants and Toddlers Longitudinal Face Image Database.}
    \label{fig:example}
\end{figure*}

%%%%%%%%% BODY TEXT
\section{Introduction}
Ensuring the accurate recognition of infants and toddlers is a critical issue faced by hospitals and healthcare systems worldwide. Applications like vaccine tracking, refugees, missing children, and human trafficking are among the most pressing concerns, highlighting the need for secure and reliable recognition mechanisms \cite{c1}, \cite{c2}, \cite{c3}. Traditional approaches, such as ID bracelets and ink-based footprinting, have been widely adopted in hospitals. However, these methods suffer from significant limitations, including susceptibility to human error, ease of tampering, and inconsistency in data collection \cite{c3}. In the context of rapidly evolving smart city environments, where identity services are increasingly integrated into healthcare, public safety, and citizen welfare systems, these limitations become more pronounced.

Biometric techniques, which have been successfully deployed in adult populations, offer a promising alternative for young children’s recognition. Fingerprints, palmprints, and iris patterns are some of the commonly studied modalities in biometrics. However, these approaches face considerable challenges in the context of young children. For instance, capturing reliable fingerprints or palmprint data is complicated by the small size of young children’s features and their inability to remain still during the process. Similarly, iris recognition, while highly accurate in adults, is not feasible for young children their difficulty in cooperating with imaging devices \cite{c3}. These limitations have motivated increased interest in exploring face recognition as a potential solution.

Face recognition offers several advantages for young children’s verification. It is non-invasive, cost-effective, and requires minimal user cooperation, making it suitable for large-scale deployment in hospital settings \cite{c4}. While face recognition has been extensively studied for adults, its application to young children is relatively nascent due to several unique challenges. Young children’s facial features are highly dynamic, undergoing rapid changes in morphology during the first few years of life. Additionally, pose, expression, and illumination variations further complicate the recognition process, as young children are uncooperative subjects who are often crying, moving, or sleeping during image acquisition \cite{c3}. Despite these challenges, preliminary studies have demonstrated the potential of face recognition in this context. For instance, techniques combining Speeded Up Robust Features (SURF) and Local Binary Patterns (LBP) have achieved promising results, with rank-1 recognition accuracy reaching up to 86.9\% on small newborn databases \cite{c2}. Similarly, hybrid models integrating Hidden Markov Models (HMM) with Singular Value Decomposition (SVD) coefficients have shown robust performance by encoding discriminative features across different scales \cite{c4}. Building on these efforts, Rowden et al. \cite{c1} introduce a longitudinal evaluation framework that systematically analyzes the impact of facial growth on recognition performance across different developmental stages. They study a commercial-off-the-shelf (COTS) algorithm on a newly of newborns, infants, and toddlers, collected dataset over a period of one year, highlights the challenges of long-term face recognition due to age progression, pose variation, and environmental inconsistencies. Unlike prior works that focus on short-term recognition \cite{c2, c4}, their findings emphasize the need for adaptive face recognition models capable of maintaining reliable identity verification as a child’s facial structure evolves over time.

However, a significant gap in literature remains. Most existing studies on young children face recognition focus on static datasets, with limited exploration of how developmental changes impact recognition accuracy over time. The rapid morphological changes that occur during infancy raise an important question: how can face recognition systems adapt to the dynamic nature of facial features across developmental stages? Addressing this question requires a longitudinal approach which captures the temporal variations in facial morphology in order to evaluate the limitations and potential of existing algorithms \cite{c1}, \cite{c4}.

This study aims to address this gap by performing a longitudinal analysis of deep learning-based face recognition systems for infants and toddlers. By systematically examining the effects of aging on facial features across developmental stages, this research seeks to establish the foundation for age-adaptive face recognition frameworks. Compared to prior work by Rowden et al. \cite{c1}, we focused on modern deep learning-based open source face recognition models for infants and toddlers, with a particular emphasis on verification accuracy over time using a novel longitudinal dataset. Our dataset consists of face images of young children aged 0 to 3 years, collected over a longitudinal period of approximately 24 months across multiple recording sessions. The longitudinal design of this study enables us to examine the impact of facial appearance changes on recognition performance as children grow. To deepen our analysis, we have partitioned the dataset into distinct age groups, allowing us to investigate patterns in recognition accuracy across demographic variables. Specifically, we aim to identify whether recognition accuracy remains consistent across age groups or whether certain groups exhibit significant declines in performance. Our contributions are summarized as follows:

\begin{itemize}
    \item \textbf{Study of Longitudinal Performance for infants and toddlers less than 3 years old:}  
    We evaluate the performance of face recognition systems as the time gap of up to 24 months between enrollment and verification samples increases. This evaluation extends over a broad age range, from infants to toddlers (ages 0–3 years), and provides insights into how aging affects the reliability of face recognition systems during this critical developmental period. The findings are directly relevant to practical scenarios—such as pediatric healthcare, vaccination tracking, and missing child recognition—where secure and accurate recognition is vital. These applications reflect the needs of emerging smart city infrastructures that rely on trustworthy, responsive biometric services for vulnerable populations.

    \item \textbf{Performance Analysis Across Age Groups:} 
    We conduct a detailed analysis of face recognition performance for children aged 0 to 3 years, using a longitudinal dataset collected across multiple sessions. The analysis highlights how recognition accuracy varies with age and time, offering insights into the reliability and limitations of current deep learning-based identity verification methods in early childhood. The results inform the design of robust, age-aware biometric systems in smart city domains such as public health, early education, and digital safety, where identity verification must adapt to a child’s rapid growth and ensure privacy.

    \item \textbf{Domain-Adversarial Adaptation for Temporal Invariance:} 
    To mitigate the effects of temporal embedding drift in young children's face data, we apply Domain-Adversarial Neural Network (DANN) approach. This technique enhances the temporal consistency of face embeddings by reducing session-specific variability while preserving identity information. This adaptation yielded significant improvement in verification performance over time, supporting the development of biometric authentication systems that can adapt to physiological changes. Such models are well-suited for deployment in privacy-conscious or decentralized smart city environments where biometric data may be processed locally or on edge devices.

    \item \textbf{Collection of a Novel Longitudinal Database:}  
    This research is grounded in the collection of the \textit{Infants and Toddlers Longitudinal Face (ITLF)} dataset, which includes high-quality face images of children aged 0 to 3 years, acquired across seven sessions over a 24-month period. The dataset enables a unique longitudinal study of biometric performance across early developmental stages. Its structure supports future investigations into building identity systems that remain effective despite rapid physiological changes and limited cooperation from young subjects. It offers a strong foundation for developing continuous and privacy-aware authentication systems in sensitive contexts such as hospitals, early childhood programs, and smart city safety networks.
\end{itemize}
%%%%%%%%%%%%%%%%%%%%%%%%%%%%%%%%%%%%%%%%%%%%%%%%%%%%%%%%%%%%%%%%%%%%%%%%%%%%%%%%

\section{Related Work}

Biometric-based recognition systems have demonstrated significant advancements in recent years, with modalities such as fingerprints, palmprints, iris recognition, and face recognition achieving high accuracy in adults, including for liveness detection and post-mortem identification \cite{b24}, \cite{b25}, \cite{b26}, \cite{b27}, \cite{b28}. However, the application of these systems to infants and toddlers remains underexplored and faces unique challenges due to their physical and behavioral characteristics. These limitations are further compounded when designing biometric identity solutions for smart city infrastructures, where secure and scalable authentication must account for age-specific constraints without relying on centralised systems.

Traditional approaches to young child recognition, such as ID bracelets and footprinting, are widely used in hospital settings. Footprinting is employed by over 90\% of hospitals in the United States, where the footprints of newborns are collected alongside the mother's fingerprints within the first two hours of birth \cite{c5}. However, these methods have well-documented limitations. Shepard et al. \cite{c6} evaluated footprints collected from 51 newborns and found that only 10 could be correctly identified by trained experts. Similarly, Pela et al. \cite{c7} analyzed 1,917 footprints collected in Brazilian hospitals and concluded that none provided sufficient detail for reliable recognition. The susceptibility of ink-based footprints to smudging and the inconsistency of data collection further limit their practicality, especially outside hospital environments \cite{c5},\cite{c7}. Again, fingerprints and palmprints are widely used for adult biometric recognition due to the permanence of ridge patterns, but their application in young children is challenging due to small ridge structures and uncooperative behavior. Weingaertner et al. \cite{c9} explored high-resolution sensors for capturing infant palmprints and footprints, achieving recognition accuracies of 83\% and 67.7\%, respectively, though the high cost and difficulty in obtaining quality ridge patterns limit practical use. Similarly, while iris recognition is highly accurate in adults \cite{c10}, there has been very little research into the effectiveness for young children, \cite{c11}.

Face recognition has emerged as a possible solution for young children verification, offering a non-invasive, cost-effective approach that requires minimal user cooperation. Face recognition offers a natural and intuitive means of identifying individuals and can complement other modalities. Recent advancements in deep learning-based methods have significantly advanced the field of biometrics. However, applying these techniques to young children presents unique challenges due to the dynamic nature of their facial features. Rapid morphological changes during early developmental stages, along with variations in pose, illumination, and expression, can significantly impact the performance of these systems \cite{c1}. In decentralised environments, the adaptive nature of facial features in infants calls for real-time and privacy-preserving learning strategies that avoid centralised repositories and allow localised updates to identity templates.

Initial efforts to apply face recognition to young children have laid the ground work in this era. Bharadwaj et al. \cite{c2} introduced a multiresolution algorithm combining Speeded Up Robust Features (SURF) and Local Binary Patterns (LBP). Their method achieved a rank-1 recognition accuracy of 86.9\% on a dataset of 34 newborns, demonstrating the potential of face recognition for infant recognition. Similarly, Goyal et al. \cite{c4} proposed a hybrid approach integrating Hidden Markov Models (HMM) with Singular Value Decomposition (SVD) coefficients. This method encoded discriminative features at multiple scales, showing robust performance even under varying pose and illumination conditions \cite{c4}. Despite the promising advancements in infant face recognition, none of these studies have explored the challenges associated with longitudinal recognition, where face images are captured over an extended period to assess the impact of facial growth on recognition performance. The absence of longitudinal evaluation leaves a critical gap in understanding how well face recognition models can maintain identity consistency as a child’s facial features evolve. Addressing this gap is essential for developing robust, age-aware biometric systems that can reliably track identity over time in real-world applications. 

While the potential of face recognition for young children is evident, significant gaps remain in the current research. Most studies rely on single session, with limited attention to the longitudinal impact of aging on facial features. Young children undergo rapid and non-linear changes in facial morphology, raising critical questions about the robustness of face recognition systems over time \cite{c1}. Moreover, the acquisition of high-quality facial images is complicated by young children’s uncooperative behavior, such as crying, movement, and sleeping during the imaging process \cite{c11}. Unlike prior studies that focus on single session or short term datasets for young children face recognition, Rowden et al. \cite{c1} introduce a longitudinal evaluation framework, capturing face images of newborns, infants, and toddlers over multiple sessions spanning an extended period using a Commercial Off-The-Shelf (COTS) face recognition algorithm available at that time. This study systematically examines how facial growth affects recognition performance over time, offering insights into the temporal stability of face embeddings in young children. By utilizing a newly collected longitudinal face image dataset, Rowden et al. \cite{c1} analyze face recognition performance across different developmental stages, addressing key challenges such as age progression, pose variation, and environmental inconsistencies. Their work contrasts with earlier methods, such as those by Bharadwaj et al. \cite{c2} and Goyal et al. \cite{c4}, which focus on short-term recognition without considering the impact of long-term morphological changes. The longitudinal approach presented in this study is essential for developing adaptive face recognition models that maintain reliable identity verification as a child’s facial structure evolves over time. Incorporating such longitudinal evaluation strategies into decentralised smart city frameworks would support identity consistency across domains such as education, healthcare, and child protection services while minimizing risk associated with centralised data breach scenarios.

Although Rowden et al. introduced a longitudinal study for infant face recognition, the field of deep learning-based face recognition has since undergone significant advancements. Over the past few years, several high-performing deep learning models, such as FaceNet, ArcFace, MagFace, and CosFace, have been developed and have demonstrated state-of-the-art performance in face recognition tasks. These models use deep convolutional neural networks (CNNs) and advanced loss functions to learn highly discriminative and robust facial embeddings, significantly improving recognition accuracy and generalization across diverse datasets. While Rowden et al.'s study provided a foundational framework for analyzing longitudinal face recognition in young children, it predates the development of these modern deep learning architectures. Given the superior performance of these models in general face recognition tasks, there is a strong motivation to evaluate their effectiveness in the challenging domain of infant and toddler recognition. The inclusion of such deep models in decentralised environments may benefit from training paradigms like federated learning, enabling multiple institutions or regions within a smart city to collaborate without sharing sensitive face data.

Building on these developments, this study aims to assess the feasibility of these state-of-the-art models in a longitudinal setting, determining whether deep learning-based approaches can overcome the inherent difficulties posed by age progression, morphological changes, and high inter-class similarity in early childhood. By benchmarking these modern open-source models against a carefully curated longitudinal dataset, we seek to provide valuable insights into their reliability, robustness, and potential applicability for real-world infant identification and verification systems. While achieving the high accuracy required for controlled environments, such as healthcare settings, may be difficult, face recognition can still serve as a valuable investigative tool. For instance, in law enforcement applications, it could assist in identifying missing children within a semi-automatic "human-in-the-loop" framework, where human oversight complements algorithmic predictions. Such applications highlight the necessity of evaluating system performance, identifying challenges, and understanding how to optimize face recognition technology for real-world scenarios. By bridging the gap between academic research and practical deployment, this study aims to contribute toward developing robust, age-aware biometric systems that can effectively support longitudinal infant and toddler face recognition in applied settings.
%%%%%%%%%%%%%%%%%%%%%%%%%%%%%%%%%%%%%%%%%%%%%%%%%%%%%%%%%%%%%%%%%%%%%%%%%%%%%%%%

\section{Longitudinal Data Collection}

Our longitudinal face data collection was conducted as part of a study investigating the feasibility of using face to reliably recognize young children ages 0-3 years old. Data were collected across seven sessions conducted at our laboratory and a local hospital. Subjects were enrolled through voluntary participation, with informed consent obtained from the parents or legal guardians of the young children, in accordance with an approved IRB protocol. The data collection process was facilitated by trained students from our lab, who operated the data collection stations at each location. This paper specifically focuses on analyzing the longitudinal face image database derived from this data collection effort.

Figure 1 presents example images of young children captured over multiple sessions as part of our longitudinal data collection using the Infants and Toddlers Longitudinal Face (ITLF) dataset. The dataset comprises 630 face images from 30 subjects, collected across seven sessions over approximately a 24-month period. Most subjects are of Caucasian background, however, ITLF does not provide
self-reported ethnicity labels. ITLF is balanced in terms of gender. Figure 2 illustrates the distribution of images across different age groups and their respective percentages. The longitudinal design of ITLF enables focused analysis of age progression and the effects of facial changes over time, particularly during the critical developmental window from 0 to 3 years. While the number of subjects is relatively small compared to some publicly available datasets, its strength lies in capturing multiple face images of the same individuals across seven sessions over 2 years. This structure allows for detailed evaluation of face recognition performance as the time gap between enrollment and verification increases, directly addressing challenges posed by rapid morphological changes during early childhood. Furthermore, despite the modest cohort size, the ITLF dataset remains a valid benchmark for evaluation, as prior studies have successfully assessed algorithms using a similar number of young children as subjects \cite{c2}.

Face images were captured with a Canon EOS 90D DSLR camera under controlled conditions to ensure high-quality data and consistency. The camera setup and image capture process were standardized as follows:

\subsection{Camera Placement and Setup}
\begin{itemize}
    \item The children were seated on a chair during the image capture. For very young children who could not sit independently, parents held the child on their lap to facilitate proper positioning.
    \item The camera was not mounted on a tripod; instead, images were captured manually to ensure optimal framing and flexibility in accommodating the child’s positioning.
    \item The background was left natural, resulting in non-uniform backgrounds in the captured images.
    \item The chair was positioned at an appropriate distance from the camera to ensure that the full face of the child was seen. Adjustments were made by moving the camera, not the subject, to preserve the child’s natural posture.
\end{itemize}

\subsection{Camera and Software Settings}
To ensure optimal image quality, the following camera settings and software configurations were used:
\begin{itemize}
    \item Camera Settings:
    \begin{itemize}
        \item \textbf{Trigger:} Remote or Manual
        \item \textbf{Flash:} Off
        \item \textbf{Lens:} Fully rotated clockwise
        \item \textbf{Image Quality:} L (best quality)
        \item \textbf{Focus:} Auto
        \item \textbf{Image Storage:} Captured images were transferred via USB cable from the camera to a connected computer for immediate storage and organization.
    \end{itemize}
    \item Software Configuration:
    \begin{itemize}
        \item The EOS Utility Software was used for remote shooting and image capture.
        \item The software preferences were configured to enable camera setting adjustments and seamless remote image capture, which streamlines the data collection workflow.
    \end{itemize}
\end{itemize}

\subsection{Collection Process and Conditions}
During each visit, multiple face images were captured from each subject to ensure high-quality data. The images were largely unconstrained, with natural variations in pose, expression, and cooperation levels, as young subjects were often uncooperative. The data collection process was designed to reflect real-world conditions. Images were captured quickly with minimal adjustments, even when subjects were crying or distressed, making the process particularly challenging.

To organize the longitudinal data, we defined Session 1 as the subject’s first visit, during which the initial face images were captured. Subsequent sessions were determined based on the time elapsed since that initial visit:

\begin{itemize}
    \item Session 2 includes images captured 0–4 months after Session 1,

    \item Session 3 includes images captured 4–8 months later,

    \item and so on,

    \item with Session 7 comprising images captured 20–24 months after the initial data collection.
\end{itemize}

This session-based categorization was necessary due to the nature of the dataset collection, which was coordinated around the subjects’ scheduled appointments with their doctors at the hospital. This structure allowed us to maintain a consistent and meaningful longitudinal organization of the dataset.

%%%%%%%%%%%%%%%%%%%%%%%%%%%%%%%%%%%%%%%%%%%%%%%%%%%%%%%%%%%%%%%%%%%%%%%%%%%%%%%%

\section{Methodology}

The primary objective of this research is to develop and evaluate a longitudinal age enrollment and verification system. The methodology consists of two primary stages: enrollment and verification.

\begin{itemize}
    \item \textbf{Enrollment Stage:} Any image captured during the subject's first visit (Session 1) is designated as the enrollment sample. Subjects are categorized into specific age brackets based on the time of image capture. This initial categorization forms the foundation for subsequent longitudinal evaluations.

    \item \textbf{Verification Stage:} Following the enrollment stage (Sessions 2 to 7), subsequent images of the same subjects from the longitudinal dataset are utilized as verification samples. These verification samples are captured at increasing time intervals as the subjects age, allowing for a detailed analysis of the system’s performance over time. This approach enables the assessment of the system's robustness to facial aging and changes in appearance across developmental stages.
\end{itemize}

%%%%%%%%%%%%%%%%%%%%%%%%%%%%%%%%
\begin{figure}[!t]
    \centering
    \includegraphics[width=0.9\linewidth]{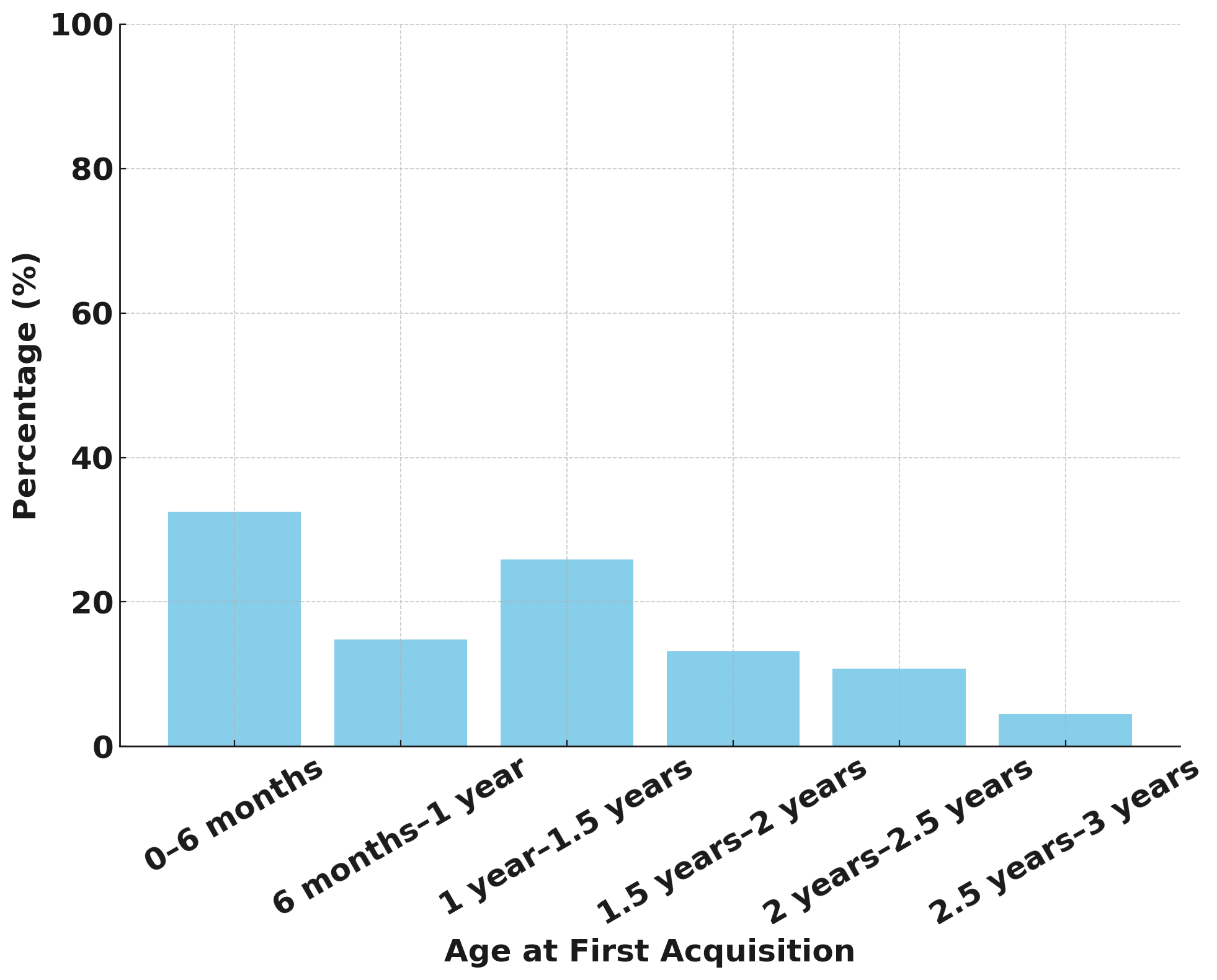}
    \caption{Distribution of collected images by age groups and their respective percentages.}
    \label{fig:age_distribution}
\end{figure}
%%%%%%%%%%%%%%%%%%%%%%%%%%%%%%%%%%%%%%%%%%

\subsection{Face Detection and Recognition}
To ensure accurate face detection and alignment, the RetinaFace, as described by Deng et al. \cite{c17}, was employed. We also analyzed the performance of other face detectors, however, RetinaFace provided better detection accuracy and alignment quality on our dataset, making it the most suitable choice for our longitudinal face recognition study. Multiple FR models were used for evaluation, including FaceNet \cite{c18}, ArcFace \cite{c19}, MagFace \cite{c20}, and CosFace \cite{c21}.

\subsection{Time-Invariant Embedding Learning via Domain Adaptation}

To address the variability in infant face representations across time, we implemented a domain-adversarial training framework based on Domain-Adversarial Neural Networks (DANN) using MagFace embeddings as input, since MagFace consistently yielded the best baseline performance on our dataset. The architecture consists of a shared feature encoder followed by two parallel branches: a domain classifier to distinguish between session-specific domains and an identity classifier to preserve subject-level discriminability. A gradient reversal layer (GRL) was integrated before the domain branch to encourage the learning of features that are invariant to temporal changes while retaining identity information. The model was trained using 80\% of the subjects from Sessions 1 through 7, with session indices as domain labels and subject IDs as identity labels, and evaluated on the remaining 20\% of subjects. This approach enabled the learning of temporally normalized embeddings, improving recognition robustness across developmental stages.

%%%%%%%%%%%%%%%%%%%%%%%%%%%%%%%%%%%%%%%%%%%%%%%%%%%%%%%%%%%%%%%%%%%%%%%%%%%%%%%%

\section{Results and Discussion}
In this work, face features were generated using four face recognition models: FaceNet, CosFace, ArcFace, and MagFace. Prior to passing the face images to these recognition models, face detection and alignment were performed using the RetinaFAce.

%%%%%%%%%%%%%%%%%%%%%%%%%%%%
\subsection{Longitudinal Evaluation of Face Recognition Performance in Infants, and Toddlers}

%%%%%%%%%%%%%%%%%%%%%%%%%%%%
\subsubsection{Performance Evaluation for Intra-Session Face Recognition}
We first evaluated the system's performance on face images captured within the same session to establish a baseline for intra-session performance. The results demonstrated that the True Acceptance Rates (TARs) exceeded 90\% at a False Acceptance Rate (FAR) of 0.1\% for all sessions (Figure 3). This high performance is expected, as intra-session variability is minimal. Face images captured within the same session of the same subject are sometime near-duplicates, taken consecutively within a short time, resulting in high similarity in their feature representations.

%%%%%%%%%%%%%%%%%%%%%%%%%%%%

\subsubsection{Impact of Temporal Aging on Face Recognition Performance}

%%%%%%%%%%%%%%%%%%%%%%%%%%%%
\begin{figure}[htbp]
    \centering
    \includegraphics[width=0.48\textwidth]{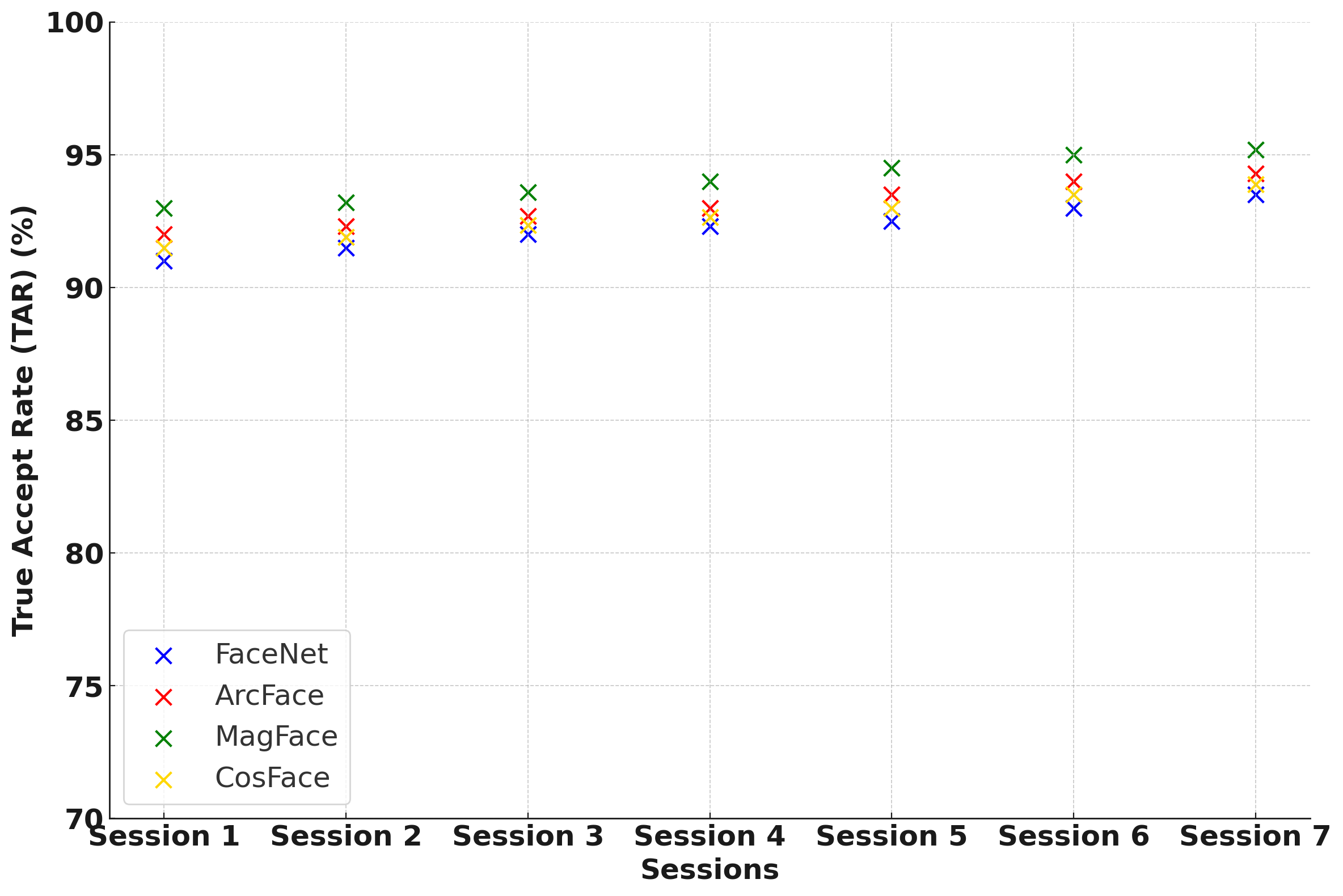} 
    \caption{True Acceptance Rate (TAR) of the FaceNet, ArcFace, MagFace, and CosFace models at a False Acceptance Rate (FAR) of 0.1\% within a session.}
    \label{fig:accuracy_plot}
\end{figure}

%%%%%%%%%%%%%%%%%%%%%%%%%
\begin{figure}[htbp]
    \centering
    \includegraphics[width=0.48\textwidth]{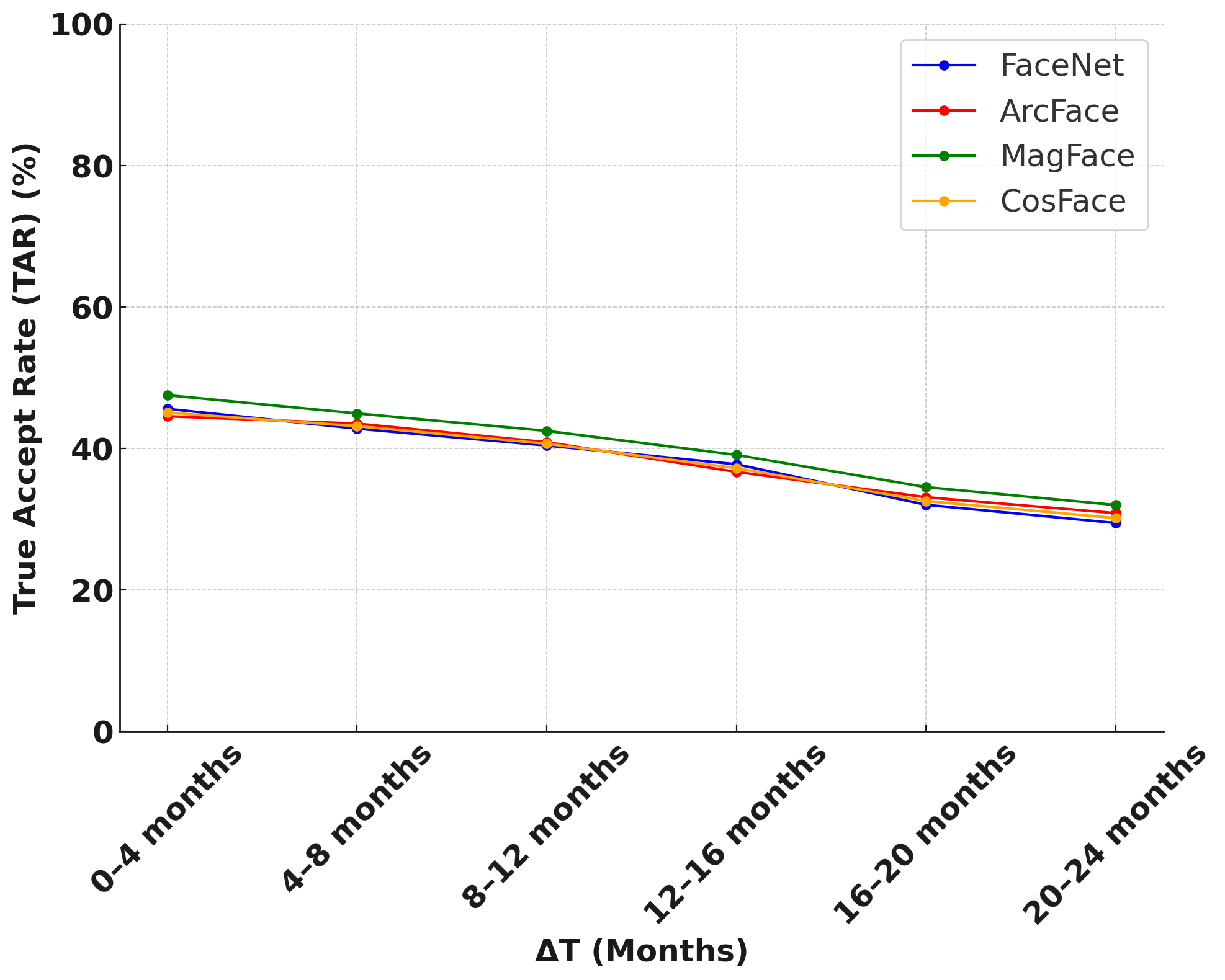} 
    \caption{True Acceptance Rate (TAR) of FaceNet, ArcFace, MagFace, and CosFace at a False Acceptance Rate (FAR) of 0.1\%, comparing initial captures with subsequent images collected over varying time intervals.}
    \label{fig:accuracy_plots_2}
\end{figure}
%%%%%%%%%%%%%%%%%%%%%%%%%%%%

To assess the longitudinal performance of the face recognition system, we conducted a comparison between face images captured during Session 1 (enrollment) and those collected during subsequent sessions (Sessions 2 through 7). This experimental setup enables us to evaluate how the system's verification accuracy evolves over an extended period of time, specifically over a 2-year elapsed time since enrollment. Figure 4 illustrates the verification performance trends across the sessions. As expected, the best performance is observed during Session 2, with a True Acceptance Rate (TAR) of 47.55\% at a False Acceptance Rate (FAR) of 0.1\% for MagFace face matcher. However, despite this relatively better performance, the system's accuracy remains quite poor, reflecting the challenges posed by rapid morphological changes in young children. Interestingly, the performance results for Session 2 and Session 3 are similar. This finding suggests that short-term aging does not introduce significant additional challenges to the system's verification performance. However, as the temporal gap increases (e.g., in Sessions 4 through 7), the system's ability to verify identities deteriorates more sharply, highlighting the non-linear nature of facial aging during early childhood.

%%%%%%%%%%%%%%%%%%%%%%%%%%%%
\subsubsection{Age-Based Performance Analysis of Face Recognition Models}
Now we are interested to analyze the performance of face recognition models on infant and toddler face images across different age groups. The subject’s age at the time of the first capture (Session 1) is used for age-based grouping. Table 1 illustrate the accuracy of four face recognition models for each age group.

%%%%%%%%%%%%%%%%%%%%%%%
\begin{table*}[t]
\centering
\small
\caption{True Acceptance Rate (TAR) of FaceNet, ArcFace, MagFace, and CosFace at a False Acceptance Rate (FAR) of 0.1\%, analyzed over different age groups.}
\resizebox{\linewidth}{!}{%
\begin{tabular}{llcccccc}
\toprule
\textbf{$\Delta T$ Interval} & \textbf{Model} & \textbf{0--6 months} & \textbf{6 mo--1 yr} & \textbf{1.0--1.5 yr} & \textbf{1.5--2.0 yr} & \textbf{2.0--2.5 yr} & \textbf{2.5--3.0 yr} \\
\midrule
0--4 months  & FaceNet  & 28.3 & 34.0 & 41.7 & 50.2 & 57.0 & 62.5 \\
             & ArcFace  & 27.4 & 33.7 & 40.4 & 48.8 & 55.5 & 61.6 \\
             & MagFace  & 30.7 & 36.5 & 43.3 & 51.7 & 58.4 & 64.7 \\
             & CosFace  & 27.9 & 33.9 & 41.0 & 49.5 & 56.2 & 62.0 \\
\midrule
4--8 months  & FaceNet  & 25.2 & 30.2 & 39.2 & 48.1 & 55.6 & 61.5 \\
             & ArcFace  & 26.5 & 31.6 & 41.2 & 50.0 & 57.1 & 62.9 \\
             & MagFace  & 27.7 & 32.9 & 43.4 & 52.3 & 59.4 & 65.0 \\
             & CosFace  & 25.9 & 30.9 & 40.7 & 49.4 & 56.8 & 62.6 \\
\midrule
8--12 months & FaceNet  & 22.4 & 27.8 & 36.7 & 46.4 & 54.5 & 60.7 \\
             & ArcFace  & 23.3 & 28.9 & 38.5 & 47.9 & 56.0 & 61.9 \\
             & MagFace  & 25.3 & 30.3 & 41.0 & 50.5 & 58.5 & 64.1 \\
             & CosFace  & 22.9 & 28.2 & 37.6 & 47.2 & 55.4 & 61.3 \\
\midrule
12--16 months & FaceNet & 19.4 & 24.1 & 32.7 & 42.3 & 50.2 & 56.3 \\
              & ArcFace & 18.9 & 23.6 & 33.2 & 40.5 & 47.9 & 53.3 \\
              & MagFace & 20.7 & 25.4 & 34.6 & 44.1 & 52.3 & 58.1 \\
              & CosFace & 19.1 & 23.8 & 33.5 & 41.7 & 49.7 & 55.5 \\
\midrule
16--20 months & FaceNet & 15.7 & 20.4 & 29.1 & 38.4 & 45.5 & 51.1 \\
              & ArcFace & 16.5 & 21.3 & 30.4 & 39.5 & 47.1 & 52.7 \\
              & MagFace & 17.7 & 22.5 & 32.0 & 41.1 & 48.8 & 54.2 \\
              & CosFace & 16.1 & 20.7 & 30.0 & 39.1 & 46.3 & 52.0 \\
\midrule
20--24 months & FaceNet & 13.4 & 18.3 & 27.3 & 36.5 & 43.4 & 48.7 \\
              & ArcFace & 13.9 & 18.9 & 28.0 & 37.5 & 44.5 & 49.6 \\
              & MagFace & 14.9 & 19.9 & 29.5 & 39.1 & 46.1 & 51.1 \\
              & CosFace & 13.8 & 18.7 & 27.8 & 36.9 & 43.9 & 49.1 \\
\bottomrule
\end{tabular}%
}
\label{tab:longitudinal_tar}
\end{table*}
%%%%%%%%%%%%%%%%%%%%%%%%%%%%%%%%%%%%%%%

The results indicate a steady improvement in face recognition accuracy with increasing age, showing the lowest performance in the 0–6 months age group and the highest accuracy in the 2.5–3 years age group. The performance drop in the 0–6 months age group can be attributed to several key factors, including the lack of stable facial features, high similarity between different individuals, and rapid soft tissue development during early infancy. These factors collectively pose substantial challenges for deep learning-based face recognition models, as they hinder the extraction of distinct identity features, making accurate verification particularly difficult at this stage. As infants grow beyond six months, their facial structures begin to stabilize, resulting in more pronounced and stable identity-specific features. This developmental transition leads to a gradual improvement in verification accuracy across all tested models. The findings of this study suggest that face recognition for children under six months remains highly unreliable, primarily due to the underdeveloped and morphologically unstable nature of neonatal facial features. However, as children approach 2.5 to 3 years of age, their facial distinctiveness and structural stability increase, enabling substantially higher recognition accuracy. This improvement suggests that face recognition for identity verification becomes increasingly feasible at this stage.

%%%%%%%%%%%%%%%%%%%%%%%%%%%%
\subsection{Time-Invariant Embedding Learning via Domain Adaptation}
To mitigate the effects of temporal variability in infant face representations, we applied a domain-adversarial training framework based on DANN using MagFace embeddings, which provided the best baseline performance. The model was trained on 80\% of subjects from Sessions 1 to 7, using session indices and identity labels in a multi-task setup with a gradient reversal layer to learn time-invariant yet identity-discriminative features. The remaining 20\% of subjects were used for evaluation.

\begin{figure}[htbp]
    \centering
    \includegraphics[width=0.48\textwidth]{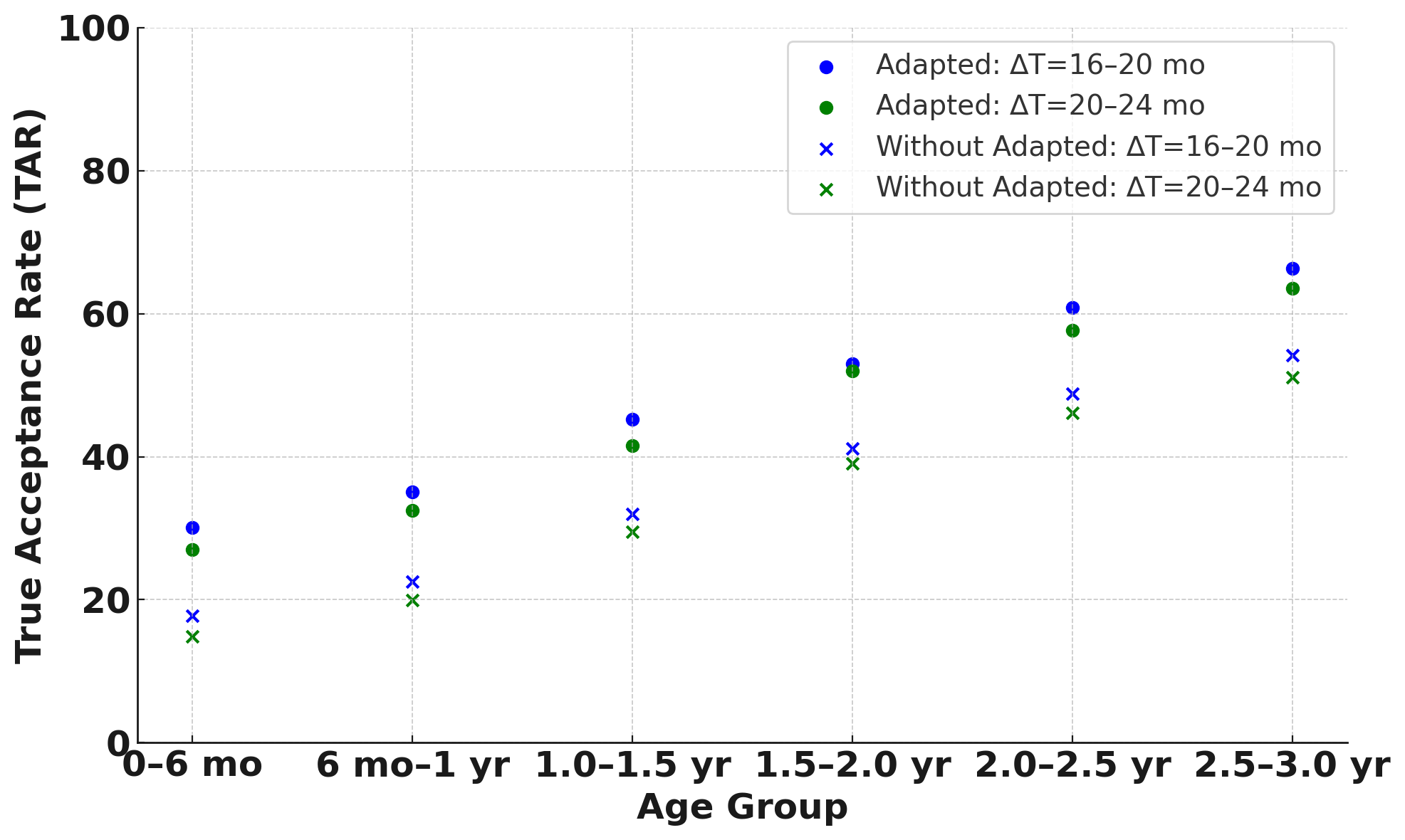} 
    \caption{True Acceptance Rate (TAR) at a False Acceptance Rate (FAR) of 0.1\% across age groups for 16–20 months and 20–24 months time intervals with and without domain-adapted features using MagFace.}
    \label{fig:accuracy_plots_2}
\end{figure}

As illustrated in Figure 5, the domain-adapted embeddings show significantly improved recognition performance when comparing Session 1 against Sessions 6 and 7, achieving TARs of 66.3\% and 63.6\%, respectively, for subjects in the 2.5–3.0 year age group. These results represent a substantial improvement over the baseline (non-adapted) performance, which achieved only 54.2\% and 51.1\% under the same conditions.
%%%%%%%%%%%%%%%%%%%%%%%%%%%%%%%%%%%%%%%%%%%%%%%%%%%%%%%%%%%%%%%%%%%%%%%%%%%%%%%%

\section{Conclusion and Future Work}
This study presents a comprehensive longitudinal evaluation of deep face recognition models for infants and toddlers, a critically underserved demographic in biometric authentication. Using a newly curated dataset of children aged 0 to 3 years captured over seven acquisition sessions, we benchmarked four widely used models—FaceNet, ArcFace, MagFace, and CosFace. Our findings underscore the significant degradation in recognition performance over time, especially among younger children, due to rapid morphological changes and identity drift, posing fundamental challenges to stable biometric authentication in early childhood.

To address the temporal instability inherent in infant facial biometrics, we employed Domain-Adversarial Neural Network (DANN) framework that adapts pretrained embeddings for improved longitudinal consistency. This approach yielded over a 12\% improvement in True Acceptance Rate (TAR) during long-term verification, demonstrating its effectiveness in mitigating session-based variability and enhancing identity persistence.

Beyond its technical contributions, this work aligns with the broader vision of secure and privacy-preserving biometric authentication in smart city ecosystems. In domains such as pediatric healthcare, child protection, and population management—where biometric identity verification must be both accurate and ethically managed—our findings reinforce the need for architectures that can adapt across time without depending on centralized repositories. Future extensions of this work include integrating multimodal biometric inputs (e.g., ear recognition) and deploying federated learning strategies to enable privacy-preserving model training across hospitals, clinics, and public health systems without compromising data sovereignty. By addressing identity verification in vulnerable populations through adaptive, decentralized-ready methods, our study contributes to the foundation for resilient biometric authentication frameworks essential for scalable, secure, and interoperable smart city services.
%%%%%%%%%%%%%%%%%%%%%%%%%%%%%%%%%%%%%%%%%%%%%%%%%%%%%%%%%%%%%%%%%%%%%%%%%%%%%%%%

\section{Acknowledgements}
This material is based upon work supported by the Center for Identification Technology Research and the National Science Foundation under Grant No. 1650503 and 2413228.

{\small
\bibliographystyle{ieee}
\bibliography{egbib}
}

\end{document}